\documentclass[10pt,conference]{IEEEtran}
\usepackage{cite}
\usepackage{amsmath,amssymb,amsfonts}
\usepackage{algorithmic}
\usepackage{algorithm}
\usepackage{graphicx}
\usepackage{textcomp}
\usepackage{xcolor}
\usepackage[T1]{fontenc}
\usepackage{xurl}
\usepackage{booktabs}
\usepackage[misc]{ifsym}
\usepackage{multirow}
\usepackage[most]{tcolorbox}
\usepackage{listings}
\usepackage{adjustbox} 
\usepackage[percent]{overpic}
\usepackage{tikz}
\usepackage{placeins}
\usepackage{stfloats}

\newtcolorbox{promptbox}[1]{%
  enhanced,
  breakable,
  enhanced jigsaw,
  colback=white,
  colframe=black!60,
  boxrule=0.4pt,
  sharp corners,
  left=1.1mm,
  right=1.1mm,
  top=0.45mm,
  bottom=0.45mm,
  before skip=3pt,
  after skip=3pt,
  colbacktitle=white,
  coltitle=black,
  fonttitle=\itshape\bfseries\footnotesize,
  center title,
  toptitle=0.25mm,
  bottomtitle=0.25mm,
  titlerule=0.3pt,
  title={#1},
}
\newtcolorbox{compactpromptbox}[1]{%
  enhanced,
  breakable,
  enhanced jigsaw,
  colback=white,
  colframe=black!85,
  boxrule=0.55pt,
  sharp corners,
  left=0.6mm,
  right=0.6mm,
  top=0.25mm,
  bottom=0.25mm,
  before skip=2pt,
  after skip=2pt,
  colbacktitle=white,
  coltitle=black,
  fonttitle=\itshape\bfseries\scriptsize,
  center title,
  toptitle=0.15mm,
  bottomtitle=0.15mm,
  titlerule=0.4pt,
  title={#1},
}
\usepackage[normalem]{ulem}

\begin{document}

\title{FactorEngine: A Program-level Knowledge-Infused Factor Mining Framework for Quantitative Investment
}

\author{
\IEEEauthorblockN{
Qinhong Lin\textsuperscript{1},
Ruitao Feng\textsuperscript{2},
Yinglun Feng\textsuperscript{1},
Zhenxin Huang\textsuperscript{3},\\
Yukun Chen\textsuperscript{1},
Zhongliang Yang\textsuperscript{1,4},
Linna Zhou\textsuperscript{1},
Binjie Fei\textsuperscript{2},
Jiaqi Liu\textsuperscript{2},
Yu Li\textsuperscript{2,*}
}
\IEEEauthorblockA{
\textsuperscript{1}Beijing University of Posts and Telecommunications, China\\
\textsuperscript{2}Beijing Value Simplex Technology Co., Ltd., China\\
\textsuperscript{3}Yangtze Delta Research Institute,
University of Electronic Science and Technology of China, China\\
\textsuperscript{4}Wuhan Eonray Technology Co., Ltd., China\\
\texttt{greenred99@bupt.edu.cn} \quad
\texttt{\{fengruitao, liyu\}@entropyreduce.com}\\
\thanks{\textsuperscript{*}Corresponding authors}
}
}

\IEEEoverridecommandlockouts
\maketitle

\begin{abstract}
We study alpha factor mining—the automated discovery of predictive signals from non-stationary market data, under a practical requirement that mined factors be directly executable and auditable, and that the discovery process remain computationally tractable at scale. 
Existing symbolic approaches search over fixed operator grammars, while neural forecasters often trade interpretability for performance and remain vulnerable to regime shifts.
We introduce FactorEngine (FE), a program-level factor discovery framework that casts factors as
execution-constrained
code and improves both effectiveness and efficiency via three separations: (i) logic revision vs. market-conditioned instantiation, (ii) LLM-guided directional search vs. Bayesian local search, and (iii) LLM usage vs. local computation. FE further incorporates a knowledge-infused bootstrapping module that transforms unstructured financial reports into executable factor programs through a closed-loop multi-agent extraction–verification–code-generation pipeline, and a trajectory-aware refinement mechanism guided by accumulated evolution experience. Across extensive backtests on real-world OHLCV data, FE produces factors with substantially stronger predictive and portfolio performance than baseline methods, including higher IC/ICIR and Rank IC/ICIR as well as improved AR and Sharpe.
\end{abstract}


\section{Introduction}
Alpha factor mining is a central objective in quantitative investment, aiming to discover predictive factors that extract actionable signals from noisy, non-stationary market data. Despite decades of research
\cite{stephens2016gplearn,hochreiter1997long,zhang2022temporal,vaswani2017attention}, effective and efficient factor discovery remains challenging due to market complexity, regime shifts, and severe noise.
\begin{figure*}[htbp]
    \centering
    \includegraphics[width=\linewidth,height=7cm]{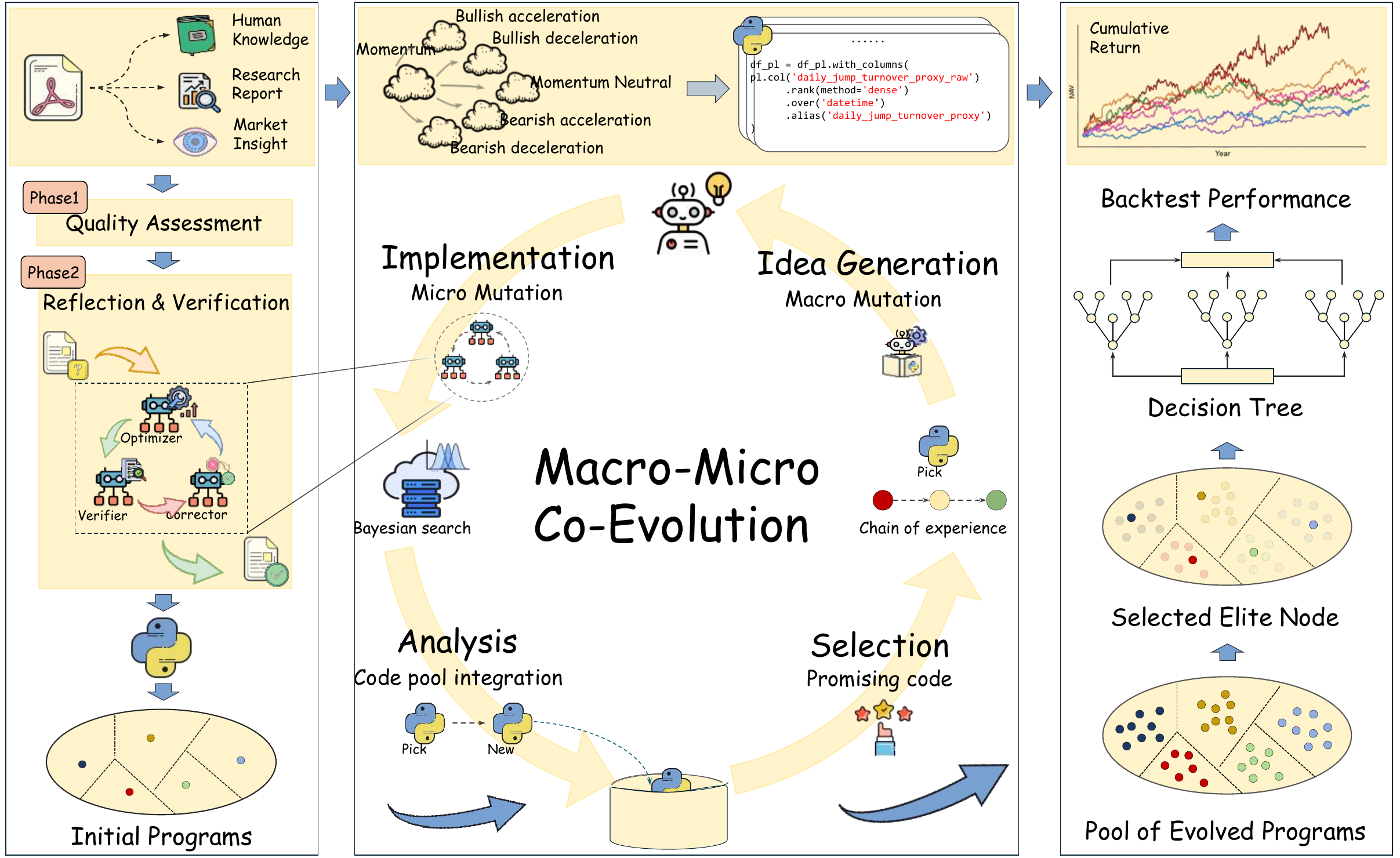} 
    \caption{\textbf{Overview of FactorEngine (FE).} \emph{Left: Bootstrapping} extracts factor ideas and converts pseudocode into executable Python to seed a knowledge-infused pool. \emph{Center: Evolution} performs macro--micro co-evolution: LLM agents propose macro mutations guided by chains of experience, and Bayesian search conducts micro-level instantiation with fast validation and feedback updates. \emph{Right: Integration} selects elite factors to train models for backtesting, producing portfolio-level feedback.
    }
    \label{fig:framework}
\end{figure*}
Existing factor mining approaches can generally be divided into two categories: symbolic expression-based methods 
and neural network-based methods. 
Symbolic factors~\cite{fama1992cross,hou2020replicating} built on explicit 
mathematical expressions provide strong interpretability and clear financial intuition while relying on
heavy manual effort and limited scalability. Additionally, symbolic factors tend to be 
fragile in the face of rapidly changing market conditions and less adaptive to real-world complexities.
Recently, genetic programming (GP) and reinforcement learning (RL) 
\cite{shi2025alphaforge,yu2023generating,zhang2020autoalpha}
have enabled automated symbolic factor search within predefined operator spaces, accelerating evolution and preserving a degree of interpretability. However, their strong dependence on manually designed operator sets constrains expressiveness, leading to limited performance and efficiency in practice.
Neural network–based
~\cite{duan2022factorvae,xu2021rest,xu2021hist} approaches capture implicit patterns and nonlinear relationships in market data. While achieving strong performance, these methods typically suffer from poor interpretability and are prone to overfitting.
More recently, large language models (LLMs) have demonstrated remarkable capabilities across a wide range of domains~\cite{brown2020gpt3,wei2022emergent}, sparking growing interest in their application to alpha mining. 
AlphaAgent~\cite{tang2025alphaagent} integrates LLM reasoning with financial report knowledge and regularized exploration to mitigate alpha decay, while RD-Agent-Quant (RD-AGENT)~\cite{li2025rdagentquant} proposes an agent-based, data-centric framework for factor and model joint optimization.

However, practical alpha mining requires that signals be not only predictive but also executable, auditable, and efficiently discoverable under practical API and wall-clock budgets. Current methods face three fundamental gaps:
\textbf{(1) Fixed symbolic grammars limit expressiveness.}
Symbolic factors~\cite{stephens2016gplearn,cui2021alphaevolve}  are searched over limited operator spaces, and new operators demand expert-designed semantics and composition rules. The expressiveness boundary is thus permanently human-defined, not search-discovered.
\textbf{(2) LLM-based methods conflate semantic search with numerical tuning.}
Recent LLM-based approaches~\cite{tang2025alphaagent,li2025rdagentquant} mix semantic mutation and market-conditioned numerical instantiation into one expensive loop. Effective factors require both computational structure and parameters encoding financial semantics: \texttt{corr5} and \texttt{corr60} share correlation logic but encode distinct regimes via window lengths. Numerical instantiation demands repeated execution over market data to sense validation fitness landscapes. The conflation wastes API budget on tasks less suited to LLM capabilities and contaminates logic signals with parameter noise.
\textbf{(3) External domain knowledge remains underutilized.} 
Financial reports contain multi-step computational logic that compact symbolic grammars do not naturally encode. Existing LLM-based methods that ingest reports often treat them as context stimulation, rather than converting report logic into executable factors.

To address these gaps, we introduce \textbf{FactorEngine (FE)}, a programmatic factor discovery framework that represents factors as executable programs satisfying explicit I/O contracts.
FE operates at two levels: at the macro level, LLMs propose parameterized logic families (computational structures with variable parameter ranges); at the micro level, Bayesian optimization determines market-conditioned instantiations (which parameter values yield economically validated factors under empirical feedback). This separation shifts expensive and slow LLM reasoning toward cheap and fast local search while preserving semantic exploration. FE further incorporates a bootstrapping module that transforms financial reports into executable seed programs, and a trajectory-aware feedback mechanism that conditions future mutations on execution history.

We evaluate FE on CSI300 and CSI500 using two configurations: \textbf{FE-alpha} (initialized with standard handcrafted factors) and \textbf{FE-report} (a knowledge-seeded variant initialized from financial reports).

Our contributions are as follows:
\begin{itemize}
    \item \textbf{Program-Level Hyper-Heuristic Framework:} We propose FactorEngine (FE), a system that transforms factor mining into a program evolution problem, extending alpha mining beyond fixed symbolic grammars.
    \item \textbf{Macro-Micro Co-evolution:} We decouple LLM-guided logic mutation from Bayesian local instantiation, alleviating local optima and efficiency bottlenecks while reducing evolution cost. 
    \item \textbf{Knowledge-Infused Factor Diversity:} We propose a closed-loop multi-agent module that operationalizes financial reports into executable seed programs, exploiting prior knowledge grounded in transparent economic rationales. 
    \item \textbf{Superior Performance \& Diversity:} Extensive experiments demonstrate that FE-alpha surpasses baselines in predictive and portfolio metrics, as well as diversity under matched information budgets. FE-report further evaluates the benefit of report-to-executable bootstrapping, achieving a 58\% improvement in Information Coefficient (IC) and a 126\% increase in excess annual return compared to Alpha158.
\end{itemize}

\section{Related Work}

\noindent \textbf{Traditional Alpha Mining}
Traditional alpha mining involves handcrafted factors derived from financial domain knowledge, such as Alpha158\footnote{https://github.com/microsoft/qlib\label{fn:qlib}} and Alpha360\footref{fn:qlib} from Qlib, which are known for their stability and powerful performance. However, manual factor design is labor-intensive and difficult to scale, which motivates automated symbolic approaches.
To this end, Genetic programming (GP) methods
automatically discover factors with predefined operators.
AlphaEvolve~\cite{cui2021alphaevolve} further enhances GP with optimization over parameters and matrix-based operations, incorporating AutoML techniques.
In parallel, reinforcement learning (RL)–based methods
formulate factor mining as a sequential decision-making problem, using financial signals such as Sharpe or Calmar ratios as rewards.
AlphaForge~\cite{shi2025alphaforge} adopts a two-stage RL framework to discover factor combinations and adaptively adjust weights.
Neural factor models~\cite{duan2022factorvae,xu2021rest,xu2021hist,li2024master}
learn latent nonlinear representations from market panels and can capture complex temporal and cross-sectional dependencies. However, their outputs are typically latent model representations rather than explicit, executable factor programs with inspectable logic and parameterized interfaces. FE complements these models by searching and validating program-level factors before their optional use in downstream predictive models.

\noindent \textbf{LLM-based Factor Discovery}
Recent work applies LLMs to alpha mining.
FAMA~\cite{li2024can} introduces dynamic factor combination and cross-sample selection to adapt across market regimes.
Shi et al.~\cite{shi2025navigating} leverages LLM-powered Monte Carlo Tree Search to improve exploration efficiency.
AlphaAgent~\cite{tang2025alphaagent} leverages agent-based frameworks to extract inspiration for factor evolution from financial materials, incorporating diversity-aware constraints to mitigate factor decay. RD-Agent~\cite{li2025rdagentquant} proposes a data-centric framework that jointly evolves factors and multi-factor models, achieving an end-to-end automation pipeline that translates model knowledge to symbolic expressions and executable code.

\noindent \textbf{Difference from prior work} Although these methods significantly improve performance, they still rely on symbolic representations, restricting expressive power and search space. Furthermore, LLMs in these works are required to handle both logic evolution and parameter optimization, limiting scalability and evolution efficiency.
Prior work primarily uses reports as textual inspiration, agent context, or symbolic-expression guidance. In contrast, FE operationalizes report knowledge into executable and testable factor programs under a unified contract. FE also separates LLM-guided logic evolution from local numerical instantiation, reducing LLM/API and wall-clock cost in our evaluated setting.

\section{Problem Formulation}
Consider an $N$-stock universe observed over $T$ trading days. Let $X_{t-L+1:t}\in \mathbb R ^{N \times L \times M}$ denote the $M$ raw market-feature channels over a lookback window of length $L$ ending at day $t$. Let $y_{t+h} \in \mathbb{R}^{N}$ denotes returns at horizon $h$.

\textbf{Mining objective.} A factor maps historical features to a cross-sectional score $f(X_{t-L+1:t})\rightarrow r_t\in\mathbb R^N$. Multiple factors $\{f_k\}$ are aggregated by a downstream model $g$ (e.g., LightGBM) for portfolio construction. Factor mining seeks programs and parameters that maximize a predictive metric on validation data:
\[
P^*, \theta^* = \arg\max_{P \in \mathcal{P},\; \theta \in \Theta(P)} J(P,\theta)
\]
where $J(P,\theta)$ can be instantiated as a single-factor validation metric (e.g., IC/RankIC) or as a multi-factor portfolio metric after aggregation by $g$.
FE instantiates $J(P,\theta)$ with the single-factor validation fitness, while portfolio-level metrics are used only for downstream evaluation.

In FE, a programmatic factor is a function $P_\theta: \mathcal{D}_{t-L+1:t} \rightarrow \mathbb{R}^N$ satisfying an execution contract: (i) fixed input schema over structured market panels, (ii) fixed output schema of stock-date-aligned scores, (iii) allowed-library constraints (e.g., pandas, numpy, polars), (iv) explicit parameter schema $\theta$, and (v) leakage-free execution (no look-ahead access). This representation is stronger than fixed symbolic grammars: it permits multi-step preprocessing, conditional logic, and parameterized subroutines while preserving comparability and auditability.
We focus on programmatic factors as the fundamental representation for alpha mining.

\section{Methodology}
As illustrated in Fig.~\ref{fig:framework}, the FE system consists of three functionally decoupled yet collaboratively interacting modules, forming a closed-loop pipeline: (1) a Bootstrapping Module, which constructs a knowledge-infused initial factor pool; (2) an Evolution Module, which performs code-level factor evolution guided by chains of experience and empirical feedback signals; and (3) an Integration Module, which supports multi-factor modeling and market data backtesting.
\subsection{Bootstrapping Module}
The Bootstrapping Module extracts, refines, and transforms expert knowledge from financial reports and expert-designed factors into programmatic factors.
\begin{figure}[t]
    \centering
    \includegraphics[width=\columnwidth,height=4cm]{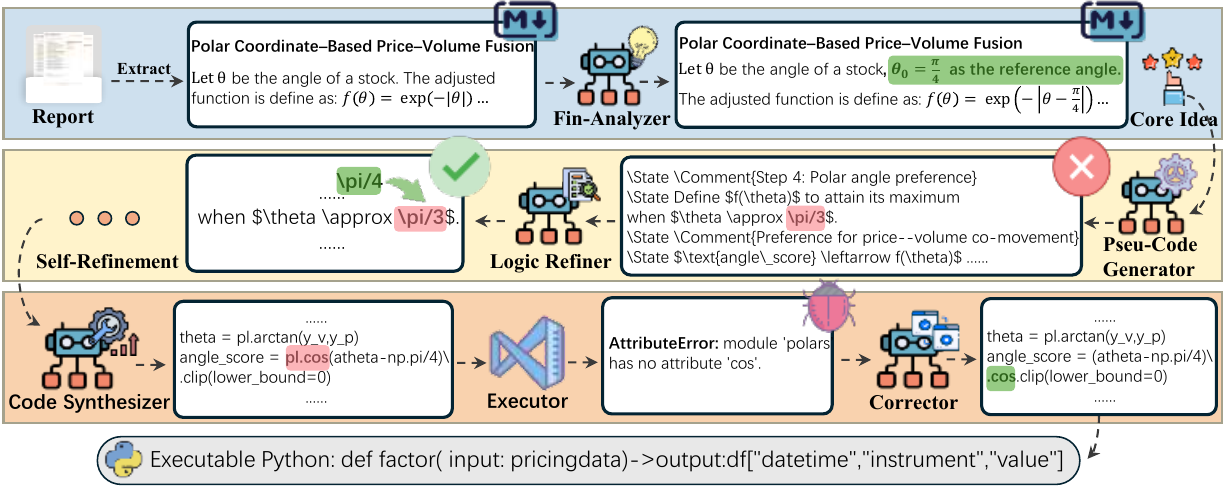} 
    \caption{Overview of the Bootstrapping module with a closed-loop extraction--verification--generation pipeline.}
    \label{fig:bootstrap}
\end{figure}
In contrast to traditional symbolic approaches that use reports only as conceptual cues for hypothesis generation, we propose a closed-loop multi-agent system that transforms 
report-derived knowledge into executable factors, thereby overcoming the limitations of predefined expression spaces.
As shown in Fig.~\ref{fig:bootstrap}, the Bootstrapping Module consists of three interconnected submodules:
\noindent (1) PDF Processing: 
Performs LLM-based compliance screening to retain valid reports, and consolidates the core knowledge with the model’s domain knowledge, yielding reliable inputs for downstream factor extraction.
\noindent(2) Factor Extraction: 
Implements a two-step understanding-to-generation workflow with iterative reflection and verification to distill core financial ideas from research reports into the structured JSON representations accompanied by LaTeX-formatted pseudocode.
\noindent (3) Code Generation: Transforms verified pseudocode and core idea 
summaries into executable Python code through iterative refinement that validates structural compliance. 

Extracted factors, along with their core idea summaries and economic rationales, form a knowledge-infused initial pool for the Evolution Module.
This module operationalizes report knowledge as executable and testable programs rather than using reports only as textual inspiration. It also identifies and repairs logical gaps before code execution, producing seed factors that can be evaluated under the same contract as expert-designed alpha factors.

\subsection{Evolution Module}
\label{sec:evolution_module}

\begin{algorithm}[t]
\small
\caption{Evolution Loop}
\label{alg:fe}
\begin{algorithmic}[1]
\REQUIRE Knowledge sources $\mathcal{K}$, iterations $R$, islands $I$, migration interval $M_{\mathrm{mig}}$
\ENSURE Elite factor set $\mathcal{F}^*$
\STATE $\mathcal{P}_0 \leftarrow \textsc{Bootstrap}(\mathcal{K})$
\STATE Initialize $I$ island trees rooted at $\mathcal{P}_0$
\FOR{$r = 1$ to $R$}
  \FOR{each island $i$ \textbf{in parallel}}
    \STATE $v \leftarrow \textsc{Select}(\text{tree}_i)$
    \STATE $C \leftarrow \textsc{BuildCoE}(v)$
    \STATE $(\Delta P, \Theta) \leftarrow \textsc{LLM-Mutate}(v, C)$ \COMMENT{Macro}
    \STATE $P' \leftarrow \textsc{Validate}(v.\text{code}, \Delta P)$
    \STATE $\theta^* \leftarrow \arg\max_{\theta \in \Theta} \text{fitness}(P', \theta)$ \COMMENT{Micro: instantiation}
    \STATE $s^* \leftarrow \textit{fitness}(P', \theta^*)$
    \STATE \textsc{AddChild}($v$, $(P', \theta^*, s^*)$); \textsc{Backpropagate}($v$, $s^*$)
  \ENDFOR
  \IF{$r \bmod M_{\mathrm{mig}} = 0$}
    \STATE \textsc{Migrate}(islands, top-3 per island)
  \ENDIF
\ENDFOR
\STATE $\mathcal{F}^* \leftarrow \{(P,\theta) : \text{fitness}(P,\theta) > \tau\}$
\RETURN $\mathcal{F}^*$
\end{algorithmic}
\end{algorithm}

The Evolution Module implements macro-micro co-evolution to address the regime-conditioned nature of alpha factors. At the macro level, LLMs propose parameterized logic families (computational templates with tunable parameter ranges), guided by trajectory-conditioned feedback. 
At the micro level, local Bayesian instantiation identifies which instantiation of each family is supported by validation data. In alpha mining, parameters such as lookback windows and decay rates encode market horizons and information decay; thus substantially different instantiations can correspond to different economic hypotheses rather than merely performance variants. This separation enables efficient exploration of both computational structure and regime-conditioned instantiations within a unified search framework.

\textbf{Full evolution loop.} As summarized in Algorithm~\ref{alg:fe}, each iteration: (1) \textit{Selection}: choose executable node by UCT~\cite{kocsis2006bandit}; (2) \textit{Mutation context}: build chain-of-experience (CoE) from current path plus diverse high-scoring paths; (3) \textit{LLM proposal}: generate a code patch and parameter search space defining a logic family; (4) \textit{Validation}: check interface compliance, leakage constraints, and executability, repair if needed; (5) \textit{Bayesian instantiation}: select the executable instance $(P,\theta)$ supported by validation data using fitness; (6) \textit{Feedback}: store best $(P,\theta,\text{fitness})$, update node statistics, summarize execution results for CoE. Periodically, top nodes are migrated across islands to share discoveries.

To guide evolution efficiently, we define a single-factor fitness metric as a surrogate objective for Bayesian search, node selection, and elite filtering:
\begin{equation}
    \text{fitness}(P,\theta) = \frac{1}{4}(10\cdot IC + ICIR + 10\cdot RIC + RICIR),
\end{equation}
which balances linear and rank-based predictive strength and stability on validation data. The sign of fitness records the direction of factor-return association, whereas its magnitude reflects predictive strength.
Elite factors selected by fitness are subsequently evaluated using portfolio-level metrics in the Integration Module.

\noindent \textbf{Program Selection.}
The evolution factor pool is organized as a tree structure, where each node corresponds to an executable program evolved from its parent and can be directly evaluated
, yielding an immediate reward equal to its validation fitness.
At each iteration, the evolution process selects the most promising node from the current tree as the context for subsequent refinement.
We define the node value $Q(v)$ as the empirical mean fitness over all evaluations within the subtree rooted at 
$v$ including itself, reflecting both its performance and its potential. 
After each evaluation, $Q(v)$ and the visit count $N(v)$ are updated via backpropagation along the ancestor path.  
To balance exploration and exploitation, we adopt the Upper Confidence Bound for Trees (UCT) criterion to score candidate nodes. Specifically, the value of a node is:
\begin{equation}
UCT(v)=Q(v)+c\sqrt{\frac{\ln N_{parent(v)}}{N_v}},    
\end{equation}
where $N_{parent(v)}$ denotes the visit count of $v$'s parent, and $c$ is a constant that controls the exploration–exploitation trade-off. We set $c = \sqrt{2}$ as it is commonly used. 
Unlike standard MCTS settings where nodes represent partial states or intermediate decisions, each node in our tree is a fully specified program and thus independently executable and evaluable, enabling flexible evolution.

\noindent \textbf{Idea Generation.}
We prompt the agent to synthesize environmental feedback and its parametric knowledge to generate high-level inspirations and structural modifications of programs.
This process emphasizes semantic reasoning, pattern abstraction, and conceptual exploration, where the LLM excels, realizing \textbf{macro mutations} as shown in Fig.~\ref{fig:framework}.

Based on the selected program node, we construct the current chain of experience (CoE) $C$ as the sequence of program nodes on the historical trajectory leading to that node. 
From the global tree, we further select $n=3$ candidate paths $\{p_i\}_{i=1}^n$
that jointly balance high empirical performance and low overlap with $C$.
Each path serves as a compact representation of prior evolution experience. 
Formally, given $C$, let $p_i = \{p_{i,1},p_{i,2},...,p_{i,|p_i|}\}$ denote a candidate path, and let $\Phi_i= p_i \cap C$ be the set of shared program nodes between $p_i$ and $C$. We define the overlap score as:
\begin{equation}
\label{eq:Scov}
    S_{cov}(p_i)= \frac{|\Phi_i|}{|p_i|}+\frac{|\Phi_i|}{|C|}
\end{equation}
which symmetrically penalizes both candidate-path overlap and current-chain overlap.
Let $P_{i,m}$ denote the program stored at node $p_{i,m}$, and let $\theta_{i,m}^*=\arg\max_{\theta\in\Theta(P_{i,m})}\text{fitness}(P_{i,m},\theta)$ be its best validation-supported instantiation.
The effectiveness score of a path is defined as the average fitness of all best-instantiated nodes on the path:
\begin{equation}
    S_{eff}(p_i)=\frac{1}{|p_i|}\sum_{m=1}^{|p_i|}
    \text{fitness}(P_{i,m},\theta_{i,m}^*).
\end{equation}
The final path score of $p_i$ is then computed as:
\begin{equation}
\label{eq:total}
  S_{total}(p_i)=S_{eff}(p_i)-S_{cov}(p_i) 
\end{equation}
These selected paths are then assembled into a structured context,
which exposes explicit experience knowledge to the LLM and cooperates with its intrinsic parametric knowledge to generate a mutation idea.
In practice, evolutionary optimization is inherently non-monotonic. Consequently, FE explicitly captures the full dynamic process, including transient fluctuations and local setbacks, rather than just the final success.
Unlike methods that condition only on static high-performing nodes, CoE conditions future mutation proposals on both successful and unsuccessful trajectories, helping the agent avoid previously degraded directions and reuse empirically useful structural changes.

\noindent \textbf{Implementation.}
In this stage, FE performs \textbf{micro mutations}: data-driven numerical instantiation of a proposed factor logic family. 
Local Bayesian instantiation employs Bayesian search, with our implementation supporting the Tree-structured Parzen Estimator (TPE). 
For a given evolved program $P$ with parameter vector $\theta \in \Theta$, FE solves $\theta^* = \arg\max_{\theta \in \Theta} \text{fitness}(P, \theta)$.
These Bayesian methods model the objective function probabilistically and suggest parameters that maximize Expected Improvement: $EI(\theta) = \int_{-\infty}^{\infty} \max(y^* - y, 0) \cdot p(y | \theta) dy$, where $y^*$ is a performance threshold (typically the top 25\% of observed 
scores), balancing exploration of uncertain regions and exploitation of 
promising regions to efficiently converge to optimal parameter combinations. During the Idea Generation phase, the LLM agent specifies parameter search ranges (e.g., window sizes, decay factors) based on domain knowledge and previous results, but the actual parameter exploration is delegated to local Bayesian search.

The validation process employs a two-phase strategy to operationalize the resource separation:
\textbf{- Phase 1}: Sequential code validation with LLM-based automatic correction, ensuring evolved programs execute correctly before parallel optimization.
\textbf{- Phase 2}: Bayesian micro-instantiation entirely on local computational resources without LLM calls. 
This phase employs multi-process parallel execution, distributing trials across multiple workers. 
The Bayesian search algorithm coordinates candidate parameters to reduce redundant exploration.
The evaluation function $\text{fitness}(P, \theta)$ computes empirical performance metrics through high-throughput execution on historical market data. 
This phase uses cached market data, multi-process execution, and efficient IC/RIC aggregation across lag periods (1, 3, 5, 10 days). This design reports local evaluations transparently while treating LLM/API calls and wall-clock time as the primary scarce resources.

\noindent \textbf{Feedback propagation.}
After a new program is instantiated and evaluated, the framework performs feedback propagation to transform empirical outcomes into actionable guidance for subsequent evolution. Specifically, the LLM is prompted to summarize the core implementation logic 
of the new program, identify its relative changes from the root and the parent program, as well as assess its performance improvement or degradation. These elements are jointly organized into a structured format and stored within the node for experience reuse in later iterations. Concurrently, the quantitative feedback is propagated back along the evolution path, updating the $Q$ and $N$ values of all traversed nodes up to the root.

Through the integration of these four stages, FE supports trajectory-conditioned mutation, macro--micro coordinated evolution, and executable validation.

Despite forming a complete evolution loop, LLM-driven mutation proposals remain stochastic. Consequently, even under identical configurations, individual evolution trajectories may diverge substantially in both structure and performance, potentially leading to inefficient exploration. 
To address this issue, we introduce a multi-island evolution configuration upon the FE framework to improve both efficiency and sampling diversity.
Specifically, we initialize $I$ independent evolution processes by duplicating the initial program node and evolve them concurrently, each of which acts as an isolated island. Every $M_{\mathrm{mig}}$ evolution rounds, each island selects its top-3 programs by fitness and migrates the copies to other islands as child nodes of the target island's root node. 
Migrated programs enter later context construction, allowing useful structural changes discovered in one trajectory to be reused by others while preserving diversity.

\subsection{Integration Module}
Building upon the bootstrapping and evolution modules, the integration module is designed to construct high-performing multi-factor signals from the evolving factor pool via elite node selection and multi-factor modeling, targeting robust predictive performance in real-world financial markets.

While explicitly modeling factor correlations with the existing factor pool is useful, the computational cost of dependency analysis grows rapidly as the factor pool expands. We therefore use a threshold-based filter: for each initial seed, we retain at most the top 5 evolved nodes whose fitness exceeds 0.4 under a rolling evaluation window of length $L=2$. For each retained node, we select the top 10 parameter configurations by the same criterion. These generated factors are then merged with Alpha158 to train the downstream LightGBM model; 
the final portfolio metrics evaluate complementary alpha rather than standalone generated-factor performance.

\begin{table*}[!b]
\centering
\scriptsize
\caption{
    Bold and underlining denote the best and second-best results, respectively; "-1"/"-2" denote 200/400 iterations. FE-alpha is the fair comparison, while FE-report uses report-derived seeds.
}
\label{tab:table_main_result}
\begin{adjustbox}{max width=\textwidth}
\begin{tabular}{l|cccccccc|cccccccc}
\toprule
\multirow{2}{*}{Methods} & \multicolumn{8}{c|}{CSI300}                                                                                                                                                    & \multicolumn{8}{c}{CSI500}                                                                                                                                                    \\
 & \multicolumn{1}{c}{IC} & \multicolumn{1}{c}{ICIR} & \multicolumn{1}{c}{RIC} & \multicolumn{1}{c}{RICIR} & \multicolumn{1}{c}{AR} & |MDD|     & IR     & \multicolumn{1}{c|}{SR} & \multicolumn{1}{c}{IC} & \multicolumn{1}{c}{ICIR} & \multicolumn{1}{c}{RIC} & \multicolumn{1}{c}{RICIR} & \multicolumn{1}{c}{AR} & |MDD|     & IR     & \multicolumn{1}{c}{SR} \\
\midrule
LGBM         & 0.0040  & 0.0326  & 0.0078  & 0.0587    & 0.0129     & 39.18\%      & 0.1706 & 0.1006 & 0.0057 & 0.0531 & 0.0108  & 0.0913    & -0.0514    & 48.22\%      & -0.3317 & -0.3051 \\
LSTM         & 0.0053  & 0.0313  & 0.0129  & 0.0704    & 0.0486     & 34.09\%      & 0.4810 & 0.3241 & 0.0054 & 0.0306 & 0.0138  & 0.0726    & -0.0104    & 36.05\%      & -0.0080 & -0.1156 \\
Transformer  & -0.0012 & -0.0066 & -0.0078 & -0.0388   & 0.0342     & 29.48\%      & 0.3027 & 0.1490 & 0.0012 & 0.0076 & -0.0024 & -0.0130   & -0.0221    & 53.04\%      & -0.0542 & -0.1598 \\
TRA          & 0.0256  & 0.1559  & 0.0302  & 0.1964    & 0.0674     & 16.02\%      & 0.6881 & 0.3747 & 0.0341 & 0.3184 & 0.0295  & 0.2717    & 0.0320     & 31.62\%      & 0.2877  & 0.0072  \\
Alpha158     & 0.0299  & 0.2008  & 0.0331  & 0.2164    & 0.0840     & 17.49\%      & 0.7440 & 0.4196 & 0.0403 & 0.3100 & 0.0416  & 0.3172    & 0.0197     & 25.17\%      & 0.2152  & 0.0089  \\
GPLearn      & 0.0292  & 0.1971  & 0.0321  & 0.2120    & 0.0814     & \underline{15.99\%}      & 0.7337 & 0.4152 & 0.0409 & \underline{0.3190} & 0.0427  & 0.3113    & 0.0272     & \underline{22.79\%}      & 0.2751  & 0.0451  \\
RD-Agent-1   & 0.0255  & 0.1667  & 0.0294  & 0.1881    & 0.0507     & 23.71\%      & 0.4770 & 0.2627 & 0.0385 & 0.2887 & 0.0398  & 0.2963    & -0.0033    & 30.55\%      & 0.0404  & -0.0915 \\
AlphaAgent-1 & 0.0282  & 0.1978  & 0.0313  & 0.2142    & 0.0673     & 17.00\%      & 0.6346 & 0.3499 & 0.0400 & 0.3076 & 0.0407  & 0.3135    & 0.0102     & 24.40\%      & 0.1431  & -0.0319 \\
FE-alpha-1   & \underline{0.0319}  & \underline{0.2178}  & \underline{0.0346}  & \underline{0.2308}    & \underline{0.0888}     & 16.91\%      & \underline{0.7886} & \underline{0.4526} & \underline{0.0413} & 0.3112 & \textbf{0.0430}  & \underline{0.3241}    & \underline{0.0346}     & \textbf{21.44\%}      & \underline{0.3315}  & \underline{0.0793}  \\
FE-report-1  & \textbf{0.0333}  & \textbf{0.2325}  & \textbf{0.0360}  & \textbf{0.2459}    & \textbf{0.1017}     & \textbf{15.89\%}      & \textbf{0.8959} & \textbf{0.5147} & \textbf{0.0420} & \textbf{0.3244} & \underline{0.0429}  & \textbf{0.3289}    & \textbf{0.0458}     & 25.44\%      & \textbf{0.4030}  & \textbf{0.1222}  \\
\midrule
RD-Agent-2   & 0.0269  & 0.1833  & 0.0300  & 0.1978    & 0.0917     & 15.23\%      & 0.8113 & 0.4647 & 0.0402 & 0.3070 & 0.0411  & 0.3130    & 0.0189     & 24.15\%      & 0.2092  & 0.0052  \\
AlphaAgent-2 & 0.0314  & 0.2089  & \underline{0.0346}  & 0.2252    & 0.0755     & 16.23\%      & 0.6779 & 0.3868 & 0.0385 & 0.2870 & 0.0396  & 0.2906    & 0.0235     & 25.36\%      & 0.2437  & 0.0266  \\
FE-alpha-2   & \underline{0.0315}  & \underline{0.2211}  & 0.0344  & \underline{0.2360}    & \underline{0.0943}     & \underline{15.07\%}      & \underline{0.8241} & \underline{0.4762} & \underline{0.0417} & \underline{0.3183} & \underline{0.0434}  & \underline{0.3293}    & \underline{0.0399}     & \underline{23.84\%}      & \underline{0.3770}  & \underline{0.1064}  \\
FE-report-2  & \textbf{0.0474}  & \textbf{0.3185}  & \textbf{0.0475}  & \textbf{0.3146}    & \textbf{0.1899}     & \textbf{12.61\%}      & \textbf{1.6001} & \textbf{1.0093} & \textbf{0.0536} & \textbf{0.4140} & \textbf{0.0487}  & \textbf{0.3744}    & \textbf{0.0836}     & \textbf{21.51\%}      & \textbf{0.6719}  & \textbf{0.2945} \\
\bottomrule
\end{tabular}
\end{adjustbox}
\end{table*}

\section{Experiment Setup}
\noindent \textbf{Baselines.} We compared FactorEngine (FE) with several representative baseline methods: (1) GPlearn~\cite{stephens2016gplearn}, which performs symbolic
factor discovery via genetic programming; (2) traditional time-series forecasting neural models, such as LightGBM (LGBM), LSTM and Transformer, which capture temporal dependencies in the market data; (3) specialized financial model, TRA~\cite{lin2021learning}, which focuses on integrating multiple trading strategies and modeling non-i.i.d. market patterns; (4) agent-based alpha factor mining methods, including AlphaAgent~\cite{tang2025alphaagent} and RD-Agent-Quant~\cite{li2025rdagentquant} (RD-Agent); (5) the hand-crafted factor baseline Alpha158.
For FE, we ran two variants. \textbf{FE-alpha} starts from expert-designed alpha factors and is the primary information-budget-matched comparison with AlphaAgent and RD-Agent. \textbf{FE-report} starts from financial reports and is a knowledge-seeded variant for evaluating report-to-executable bootstrapping. 
The report seeds are derived from Chinese sell-side quantitative research reports.
To prevent potential leakage in the knowledge-infused bootstrapping module, we only used financial research reports published before 2017 for factor extraction and code bootstrapping, ensuring that no report content overlaps with the test period.
We acknowledge that modern LLMs are trained on data extending beyond our test period. However, this limitation is shared by all agent-based methods, and we ensured fair comparison by using Gemini-2.5-Pro as the backbone model across all agent baselines, with the temperature set to 0.7.

\noindent \textbf{Hyperparameters.} 
We conducted two experiments with different budgets, in which each framework evolved 200 and 400 iterations respectively, with one factor generated per iteration. 
Under the two budgets, the FE framework was initialized with 5 and 10 alpha factors (or reports), respectively.
For FE-alpha, the 5-factor seed set is \{\texttt{corr5}, \texttt{resi5}, \texttt{klen}, \texttt{klow}, \texttt{vstd5}\}, and the 10-factor seed set is  \{\texttt{corr5}, \texttt{resi10}, \texttt{roc60}, \texttt{rsqr5}, \texttt{cord5}, \texttt{std5},  \texttt{klen}, \texttt{klow}, \texttt{vstd5}, \texttt{wvma5}\}.
For FE-report, each seed factor is derived from a single report through a conversion process.
The number of islands was set to 2, with migration performed every 7 iterations.
For each FE macro mutation, local Bayesian instantiation runs 20 trials over the LLM-proposed parameter ranges without additional LLM calls.
All generated factors were filtered according to each framework’s criteria before multi-factor modeling.
For backtesting, generated factors from all mining methods were merged with the Alpha158 set to train a LightGBM model. This setting follows common multi-factor practice and evaluates whether generated factors provide complementary alpha over a strong standard factor base.


\noindent \textbf{Trading protocol.}
Portfolio metrics are net of realistic A-share costs. The strategy ranks stocks by model 
score, buys the top 50, and uses five rolling 5-day sub-portfolios with equal weights, following the same portfolio construction and ranking scheme as all baselines. We charge commission $1.5\times10^{-4}$ on both buys and sells, stamp duty $5\times10^{-4}$ on sells, and slippage $8\times10^{-4}$ on all trades. Execution also enforces a 1-lot minimum, a 10\% daily-volume cap per stock, price-limit constraints, and initial capital of 100M CNY.
For multi-factor modeling, the LightGBM model is retrained under an ``8+1+1'' walk-forward protocol.

\noindent \textbf{Datasets.}
We used the full-market dataset, divided into training (2008-01-01 – 2014-12-31), validation (2015-01-01 – 2016-12-31), and testing (2017-01-01 – 2024-12-31) periods for factor discovery, and evaluated in the CSI300 and CSI500 markets. The raw features used for factor construction consist solely of OHLCV data.

\noindent \textbf{Metrics.}
For predictive performance, we reported the Information Coefficient (IC), Information Coefficient Information Ratio (ICIR), Rank IC, and Rank ICIR. For portfolio performance, we evaluated Annualized Return (AR),  Information Ratio (IR), maximum drawdown (MDD) and Sharpe Ratio (SR). 
Portfolio metrics are computed on the excess return series, defined as the portfolio return minus the benchmark return.

All experiments were conducted on a server equipped with 56 CPU cores, providing a total of 56 parallel threads.

\section{Results Analysis}
\subsection{Main Result}
Tab.~\ref{tab:table_main_result} presents the experimental results of FactorEngine (FE) and baseline methods in the CSI300 and CSI500 markets.
\begin{figure*}[!t]
    \centering
    \begin{minipage}[t]{0.32\textwidth}
        \centering
        \includegraphics[width=\linewidth,trim=0 8pt 0 0,clip]{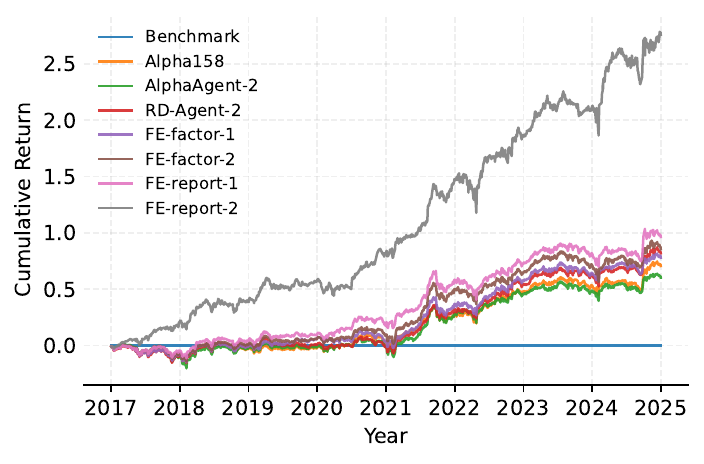}
    \end{minipage}
    \hfill
    \begin{minipage}[t]{0.32\textwidth}
        \centering
        \includegraphics[width=\linewidth,trim=0 8pt 0 0,clip]{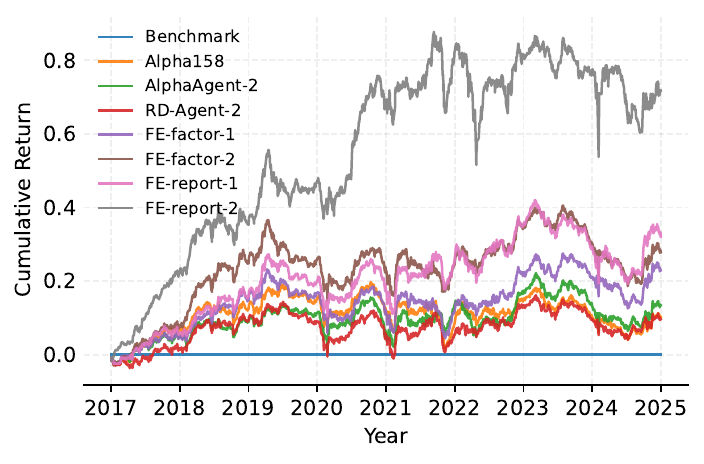}
    \end{minipage}
    \hfill
    \begin{minipage}[t]{0.24\textwidth}
        \centering
        \includegraphics[width=\linewidth,height=0.90\linewidth,trim=0 8pt 0 0,clip]{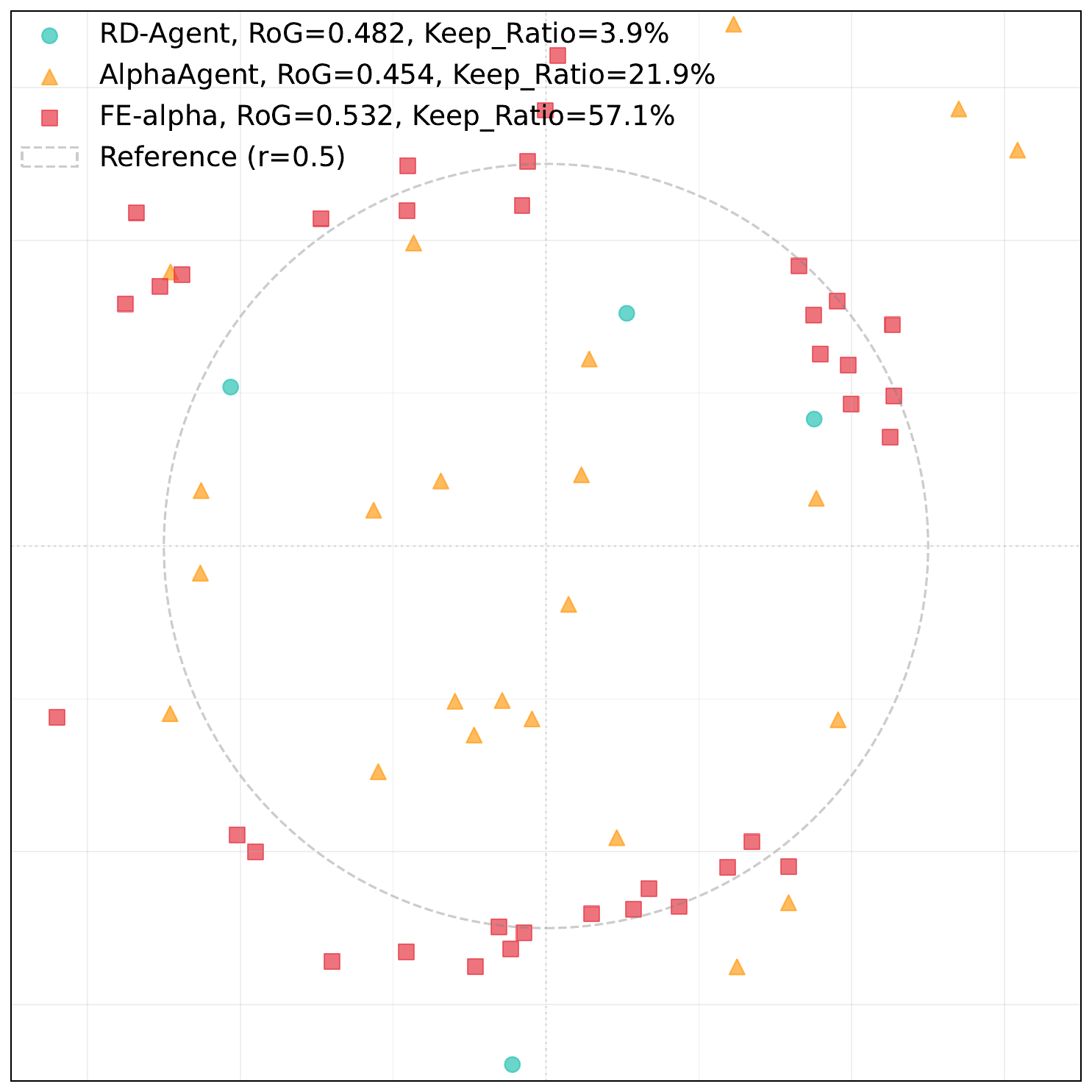} 
    \end{minipage}
    \caption{
   (Left/Middle) Net cumulative excess returns in CSI300 and CSI500 under the cost-aware trading protocol.
    (Right) Factor-pool diversity diagnostic on CSI500 via MDS on $1-|\rho|$, 
    with RoG and keep ratio reported in the legend.
    }
    \label{fig:cr_comparison}
\end{figure*}
\begin{figure*}[!b]
    \centering
    \begin{minipage}[t]{0.32\linewidth}
        \centering
        \includegraphics[width=\linewidth, height=3.5cm]{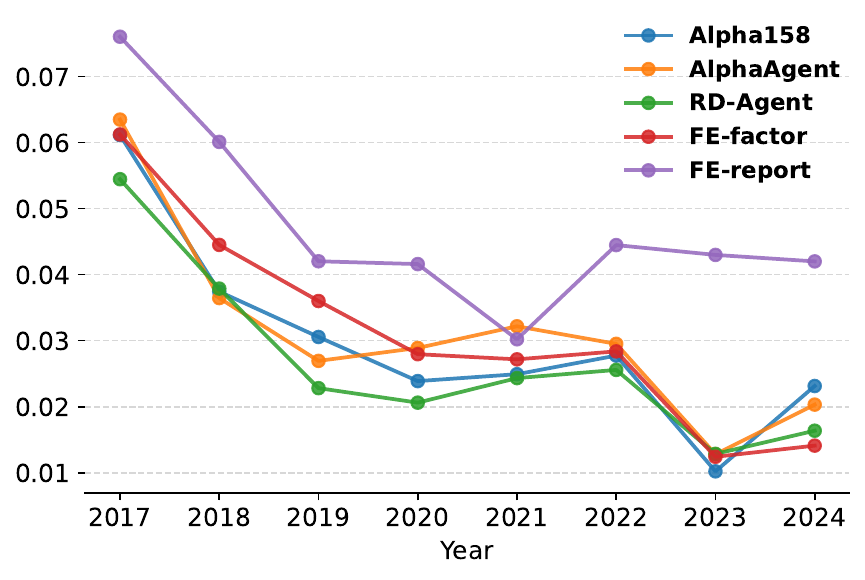}
    \end{minipage}
    \begin{minipage}[t]{0.32\linewidth}
        \centering
        \includegraphics[width=\linewidth, height=3.5cm]{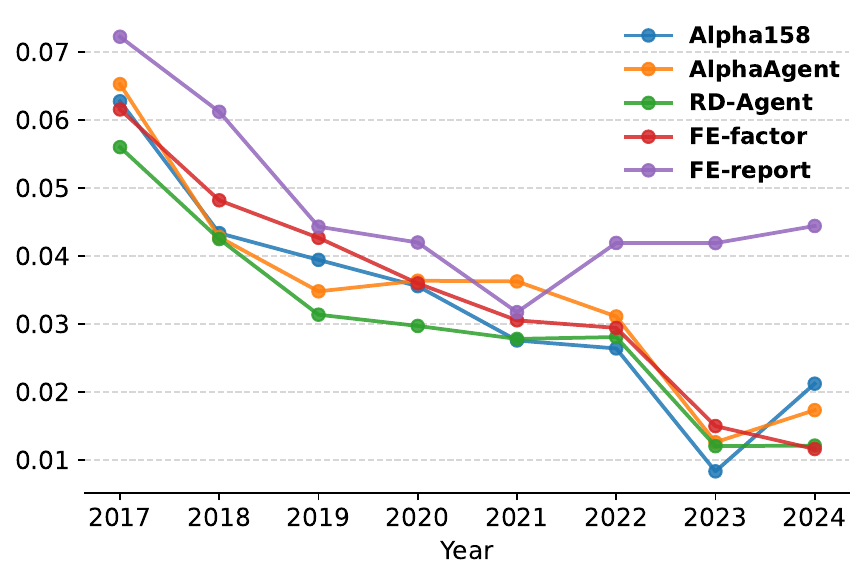}
    \end{minipage}
    \begin{minipage}[t]{0.32\linewidth}
        \centering
        \includegraphics[width=\linewidth, height=3.5cm]{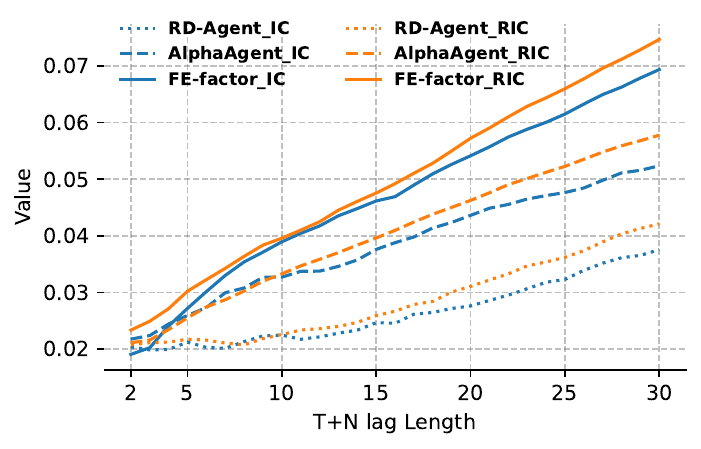}
    \end{minipage}
    \caption{Yearly IC (Left) and Rank IC (Middle) comparisons on the CSI300 market. Mean IC and Rank IC between the top 10\% factors and future returns at T+N on the CSI300 market across three settings (Right; the plot label ``T+N'' denotes an N-day-ahead return horizon).}
    \label{fig:yearly_ic_comparison}
\end{figure*}
Traditional neural models like LGBM, LSTM, and Transformer show weaker predictive and portfolio results in this setting, especially on CSI500 where their excess returns are negative. 
The TRA method used artificial factors for deep learning modeling, integrating their feature representations, which significantly enhanced the predictive capability of the RNN network.
The 50 factors generated by GPlearn show that, in the CSI300 market, mixing these 50 factors with Alpha158 does not further improve the correlation between the features and future returns. However, in the CSI500 market, GPlearn’s factors provide significant returns.
FE-alpha provides the primary fair comparison with agent-based baselines and consistently improves predictive and portfolio metrics under the same initialization budget. FE-report further improves performance when pre-2017 report-derived seeds are available, 
evaluating the end-to-end utility of report-derived initialization.
In our experiments, although three agent-based methods showed improvements as the number of iterations increased, AlphaAgent and RD-Agent performed noticeably worse than the Alpha158 factor in the CSI300 market with fewer iterations. However, as the number of iterations increased, the performance gap narrowed, and they eventually surpassed Alpha158.
In contrast, both FE setups consistently showed better predictive performance and portfolio performance than other methods at different iteration rounds. This indicates that the program-level factor evolution framework (FE) is effective at discovering factors that better capture market characteristics.
Furthermore, when the number of report-derived seed factors increased from 5 to 10, the performance of the factors evolved from financial reports significantly improved. IC increased from 0.0333 to 0.0474, AR improved from 0.1017 to 0.1899, and MDD decreased from 15.89\% to 12.61\%. It shows that FE effectively extracts financial knowledge from reports for modeling and continuously evolves factors during iterations. 

Through the two FE experimental setups, we found that the factors evolved from reports consistently outperformed those evolved from expert-designed alpha seeds, regardless of the iteration rounds. 
The results suggest that report-derived prior knowledge can provide useful initialization for factor evolution, although the separate contribution of the
bootstrapping stages is not isolated in this comparison.
Fig.~\ref{fig:cr_comparison} shows the net cumulative excess-return curves under the common cost-aware trading protocol in the CSI300 and CSI500 markets.
The factors evolved using FE outperformed Alpha158, AlphaAgent, and RD-AGENT in both 200-iteration and 400-iteration experiments. 

\subsection{Factor Diversity Analysis}
To evaluate the non-redundancy of generated factors, we first filter out low-quality factors with an IC below 0.015 and then analyze the correlation structure of the retained factor pool. We construct a dissimilarity matrix with entries $1-|\rho|$ and use Multidimensional Scaling (MDS) to embed it into a 2D space. The absolute correlation treats two factors with opposite signs but near-identical rankings as redundant, so the visualization focuses on signal novelty rather than sign direction. In the embedded space, larger distances indicate weaker absolute correlations between factors.
As shown in the right subfigure of Fig.~\ref{fig:cr_comparison}, in the CSI500 setting, FE-alpha retained 36 effective factors after quality filtering, corresponding to a 57.1\% keep ratio and significantly surpassing the yield of AlphaAgent and RD-AGENT. The legend further 
reports the Radius of Gyration (RoG), which summarizes the overall dispersion of the retained factors in the MDS space. 
Beyond the number of surviving factors, FE-alpha exhibits broader dispersion in the MDS diagnostic and the largest RoG, indicating that its retained factors are not concentrated in a narrow correlation cluster. Together, the higher keep ratio and larger dispersion suggest that FE-alpha produces a richer and less redundant factor set, offering more complementary alpha signals for downstream multi-factor models.

\subsection{Alpha Decay Analysis}
In Fig.~\ref{fig:yearly_ic_comparison}, the first two subplots show the yearly IC variation trends on the test data of the CSI300 market under the 400-iteration setting, with Alpha158 included as the handcrafted baseline. 
All factors exhibited some degree of decay over time, but throughout the entire period, FE-report consistently maintained a high IC and Rank IC. After 2021, FE-report no longer followed the earlier decay trend and showed a recovery in yearly IC/Rank IC, suggesting better adaptation to the later test-period market conditions. The FE-alpha experimental group experienced a more gradual decay compared to other factors, with no significant fluctuations, resulting in superior cumulative returns and portfolio performance.
Although the IC values of other methods were slightly higher in some years, their significant declines in other years make it difficult to mitigate the losses.
The right panel of Fig.~\ref{fig:yearly_ic_comparison} provides a single-factor performance without LightGBM aggregation.
We select the top 10\% generated factors from each agent-based method after 400 evolution rounds and compute their IC and Rank IC against forward returns at different horizons. FE consistently achieves higher IC/Rank IC across horizons, indicating that its discovered factor pool contains stronger predictive signals rather than only improving through downstream LightGBM aggregation. The stable advantage across lag lengths further suggests that FE factors are not over-specialized to a single return horizon, but capture more robust cross-sectional signals.

\subsection{Robustness of Factor Evolution}
\label{sec:evolution_robustness}
Because LLM-guided evolution is non-deterministic, final portfolio metrics alone do not reveal whether factor discovery is stable across starting conditions. We therefore analyze FE-alpha trajectories across 20 initial seed programs on CSI300 from two complementary views. Fig.~\ref{fig:fe_alpha_20_iteration_stability} illustrates how $|\mathrm{fitness}|$ varies with iteration and generation. Here, generation denotes tree depth in the evolution tree: the root seed is Gen-0, and each mutation edge increases the generation by one. For a separate stability diagnostic, we evolved 20 alpha seeds to estimate across-seed dispersion more reliably.
Unlike the mining-stage fitness used for node selection, the diagnostics in Fig.~\ref{fig:fe_alpha_20_iteration_stability} report $|\mathrm{fitness}|$ on the held-out test data.
The right plot pools observed generation depths from the multi-seed FE-alpha diagnostic; Gen-8 is the maximum observed depth in the aggregated evolution results, not a depth reached by every seed trajectory. The Case Study below is a separate single Abandoned-Baby checkpoint-40 tree whose deepest observed generation is 4.
The trajectories are non-monotonic, as expected for mutation-based search, but the descriptive statistics show an upward shift in factor quality while preserving bounded dispersion. In the right plot, the mean $|\mathrm{fitness}|$ increases from about 0.08 at Gen-0 to about 0.13 at Gen-8, while the inter-quartile range narrows from about 0.07 to 0.04. This provides factor-discovery-level robustness evidence that complements the downstream portfolio results. This analysis targets factor-discovery stability across starting factor families rather than repeated full-portfolio retraining.

\begin{figure}[!t]
    \centering
    \resizebox{\columnwidth}{!}{%
    \begin{tikzpicture}[x=1cm,y=0.8cm]
        \definecolor{feBlue}{RGB}{31,75,216}
        \definecolor{feBand}{RGB}{111,121,255}
        \definecolor{fePanel}{RGB}{238,238,246}
        \definecolor{feBox}{RGB}{214,229,255}
    \begin{scope}[shift={(0,0)}]

        \fill[fePanel] (0,0) rectangle (3.05,2.10);
        \draw[white, line width=0.32pt] (0,0.000) -- (3.05,0.000);
        \draw[white, line width=0.32pt] (0,1.050) -- (3.05,1.050);
        \draw[white, line width=0.32pt] (0,2.100) -- (3.05,2.100);
        \draw[white, line width=0.25pt] (0.000,0) -- (0.000,2.10);
        \node[below, font=\tiny] at (0.000,-0.05) {0};
        \draw[white, line width=0.25pt] (1.525,0) -- (1.525,2.10);
        \node[below, font=\tiny] at (1.525,-0.05) {20};
        \draw[white, line width=0.25pt] (3.050,0) -- (3.050,2.10);
        \node[below, font=\tiny] at (3.050,-0.05) {40};
        \draw[black!42, line width=0.28pt] (0,0) rectangle (3.05,2.10);
        \node[left, font=\tiny] at (-0.05,0.000) {0.06};
        \node[left, font=\tiny] at (-0.05,1.050) {0.09};
        \node[left, font=\tiny] at (-0.05,2.100) {0.13};
        \fill[feBand, opacity=0.24] (0.000,1.197) -- (0.076,1.048) -- (0.152,1.364) -- (0.229,1.749) -- (0.305,1.067) -- (0.381,1.479) -- (0.457,1.616) -- (0.534,1.459) -- (0.610,1.652) -- (0.686,1.477) -- (0.762,1.667) -- (0.839,1.721) -- (0.915,1.919) -- (0.991,1.811) -- (1.067,1.891) -- (1.144,1.892) -- (1.220,1.814) -- (1.296,1.616) -- (1.373,1.845) -- (1.449,1.970) -- (1.525,1.521) -- (1.601,1.753) -- (1.677,1.603) -- (1.754,1.728) -- (1.830,1.862) -- (1.906,2.020) -- (1.982,1.724) -- (2.059,1.711) -- (2.135,1.619) -- (2.211,1.688) -- (2.287,1.762) -- (2.364,1.941) -- (2.440,1.770) -- (2.516,1.801) -- (2.592,1.507) -- (2.669,1.631) -- (2.745,2.025) -- (2.821,1.725) -- (2.897,1.679) -- (2.974,1.806) -- (3.050,1.629) -- (3.050,0.500) -- (2.974,0.473) -- (2.897,0.687) -- (2.821,0.413) -- (2.745,1.230) -- (2.669,0.485) -- (2.592,0.552) -- (2.516,0.666) -- (2.440,0.619) -- (2.364,1.061) -- (2.287,0.913) -- (2.211,0.869) -- (2.135,0.621) -- (2.059,0.500) -- (1.982,0.715) -- (1.906,1.051) -- (1.830,0.841) -- (1.754,0.618) -- (1.677,0.691) -- (1.601,0.627) -- (1.525,0.673) -- (1.449,1.099) -- (1.373,0.720) -- (1.296,0.630) -- (1.220,0.695) -- (1.144,1.029) -- (1.067,0.647) -- (0.991,0.637) -- (0.915,0.720) -- (0.839,0.513) -- (0.762,0.391) -- (0.686,0.219) -- (0.610,0.503) -- (0.534,0.373) -- (0.457,0.515) -- (0.381,0.336) -- (0.305,0.066) -- (0.229,0.451) -- (0.152,0.153) -- (0.076,0.135) -- (0.000,0.035) -- cycle;
        \draw[feBlue, line width=0.70pt] (0.000,0.616) -- (0.076,0.592) -- (0.152,0.759) -- (0.229,1.100) -- (0.305,0.566) -- (0.381,0.908) -- (0.457,1.066) -- (0.534,0.916) -- (0.610,1.078) -- (0.686,0.848) -- (0.762,1.029) -- (0.839,1.117) -- (0.915,1.320) -- (0.991,1.224) -- (1.067,1.269) -- (1.144,1.461) -- (1.220,1.254) -- (1.296,1.123) -- (1.373,1.282) -- (1.449,1.535) -- (1.525,1.097) -- (1.601,1.190) -- (1.677,1.147) -- (1.754,1.173) -- (1.830,1.352) -- (1.906,1.535) -- (1.982,1.220) -- (2.059,1.105) -- (2.135,1.120) -- (2.211,1.278) -- (2.287,1.338) -- (2.364,1.501) -- (2.440,1.195) -- (2.516,1.233) -- (2.592,1.029) -- (2.669,1.058) -- (2.745,1.628) -- (2.821,1.069) -- (2.897,1.183) -- (2.974,1.140) -- (3.050,1.065);
        \fill[feBlue] (0.000,0.616) circle (0.45pt);
        \fill[feBlue] (0.152,0.759) circle (0.45pt);
        \fill[feBlue] (0.305,0.566) circle (0.45pt);
        \fill[feBlue] (0.457,1.066) circle (0.45pt);
        \fill[feBlue] (0.610,1.078) circle (0.45pt);
        \fill[feBlue] (0.762,1.029) circle (0.45pt);
        \fill[feBlue] (0.915,1.320) circle (0.45pt);
        \fill[feBlue] (1.067,1.269) circle (0.45pt);
        \fill[feBlue] (1.220,1.254) circle (0.45pt);
        \fill[feBlue] (1.373,1.282) circle (0.45pt);
        \fill[feBlue] (1.525,1.097) circle (0.45pt);
        \fill[feBlue] (1.677,1.147) circle (0.45pt);
        \fill[feBlue] (1.830,1.352) circle (0.45pt);
        \fill[feBlue] (1.982,1.220) circle (0.45pt);
        \fill[feBlue] (2.135,1.120) circle (0.45pt);
        \fill[feBlue] (2.287,1.338) circle (0.45pt);
        \fill[feBlue] (2.440,1.195) circle (0.45pt);
        \fill[feBlue] (2.592,1.029) circle (0.45pt);
        \fill[feBlue] (2.745,1.628) circle (0.45pt);
        \fill[feBlue] (2.897,1.183) circle (0.45pt);
        \fill[feBlue] (3.050,1.065) circle (0.45pt);
        \node[font=\tiny] at (1.525,-0.43) {Iteration};
        \node[font=\scriptsize, anchor=west] at (0,2.270) {Left: Iteration trajectory};
    \end{scope}
    \begin{scope}[shift={(3.770,0)}]

        \fill[fePanel] (0,0) rectangle (3.05,2.10);
        \draw[white, line width=0.32pt] (0,0.000) -- (3.05,0.000);
        \draw[white, line width=0.32pt] (0,1.050) -- (3.05,1.050);
        \draw[white, line width=0.32pt] (0,2.100) -- (3.05,2.100);
        \draw[black!42, line width=0.28pt] (0,0) rectangle (3.05,2.10);
        \node[left, font=\tiny] at (-0.05,0.000) {0.01};
        \node[left, font=\tiny] at (-0.05,1.050) {0.09};
        \node[left, font=\tiny] at (-0.05,2.100) {0.17};
        \begin{scope}[xshift=0.20cm,xscale=0.87]
        \draw[feBlue, line width=0.30pt] (0.000,0.032) -- (0.000,0.488);
        \draw[feBlue, line width=0.30pt] (0.000,1.450) -- (0.000,1.618);
        \draw[feBlue, line width=0.30pt] (-0.100,0.032) -- (0.100,0.032);
        \draw[feBlue, line width=0.30pt] (-0.100,1.618) -- (0.100,1.618);
        \filldraw[fill=feBox, draw=feBlue, line width=0.42pt] (-0.100,0.488) rectangle (0.100,1.450);
        \draw[red!72!black, line width=0.38pt] (-0.100,0.890) -- (0.100,0.890);
        \node[below, font=\tiny] at (0.000,-0.05) {0};
        \draw[feBlue, line width=0.30pt] (0.381,0.426) -- (0.381,0.701);
        \draw[feBlue, line width=0.30pt] (0.381,1.379) -- (0.381,1.606);
        \draw[feBlue, line width=0.30pt] (0.281,0.426) -- (0.481,0.426);
        \draw[feBlue, line width=0.30pt] (0.281,1.606) -- (0.481,1.606);
        \filldraw[fill=feBox, draw=feBlue, line width=0.42pt] (0.281,0.701) rectangle (0.481,1.379);
        \draw[red!72!black, line width=0.38pt] (0.281,1.012) -- (0.481,1.012);
        \node[below, font=\tiny] at (0.381,-0.05) {1};
        \draw[feBlue, line width=0.30pt] (0.762,0.409) -- (0.762,0.859);
        \draw[feBlue, line width=0.30pt] (0.762,1.465) -- (0.762,1.700);
        \draw[feBlue, line width=0.30pt] (0.662,0.409) -- (0.862,0.409);
        \draw[feBlue, line width=0.30pt] (0.662,1.700) -- (0.862,1.700);
        \filldraw[fill=feBox, draw=feBlue, line width=0.42pt] (0.662,0.859) rectangle (0.862,1.465);
        \draw[red!72!black, line width=0.38pt] (0.662,1.150) -- (0.862,1.150);
        \node[below, font=\tiny] at (0.762,-0.05) {2};
        \draw[feBlue, line width=0.30pt] (1.144,0.536) -- (1.144,0.902);
        \draw[feBlue, line width=0.30pt] (1.144,1.438) -- (1.144,1.682);
        \draw[feBlue, line width=0.30pt] (1.044,0.536) -- (1.244,0.536);
        \draw[feBlue, line width=0.30pt] (1.044,1.682) -- (1.244,1.682);
        \filldraw[fill=feBox, draw=feBlue, line width=0.42pt] (1.044,0.902) rectangle (1.244,1.438);
        \draw[red!72!black, line width=0.38pt] (1.044,1.252) -- (1.244,1.252);
        \node[below, font=\tiny] at (1.144,-0.05) {3};
        \draw[feBlue, line width=0.30pt] (1.525,0.609) -- (1.525,0.859);
        \draw[feBlue, line width=0.30pt] (1.525,1.528) -- (1.525,1.830);
        \draw[feBlue, line width=0.30pt] (1.425,0.609) -- (1.625,0.609);
        \draw[feBlue, line width=0.30pt] (1.425,1.830) -- (1.625,1.830);
        \filldraw[fill=feBox, draw=feBlue, line width=0.42pt] (1.425,0.859) rectangle (1.625,1.528);
        \draw[red!72!black, line width=0.38pt] (1.425,1.211) -- (1.625,1.211);
        \node[below, font=\tiny] at (1.525,-0.05) {4};
        \draw[feBlue, line width=0.30pt] (1.906,0.445) -- (1.906,0.853);
        \draw[feBlue, line width=0.30pt] (1.906,1.568) -- (1.906,1.824);
        \draw[feBlue, line width=0.30pt] (1.806,0.445) -- (2.006,0.445);
        \draw[feBlue, line width=0.30pt] (1.806,1.824) -- (2.006,1.824);
        \filldraw[fill=feBox, draw=feBlue, line width=0.42pt] (1.806,0.853) rectangle (2.006,1.568);
        \draw[red!72!black, line width=0.38pt] (1.806,1.209) -- (2.006,1.209);
        \node[below, font=\tiny] at (1.906,-0.05) {5};
        \draw[feBlue, line width=0.30pt] (2.287,0.003) -- (2.287,0.730);
        \draw[feBlue, line width=0.30pt] (2.287,1.525) -- (2.287,1.708);
        \draw[feBlue, line width=0.30pt] (2.187,0.003) -- (2.387,0.003);
        \draw[feBlue, line width=0.30pt] (2.187,1.708) -- (2.387,1.708);
        \filldraw[fill=feBox, draw=feBlue, line width=0.42pt] (2.187,0.730) rectangle (2.387,1.525);
        \draw[red!72!black, line width=0.38pt] (2.187,1.159) -- (2.387,1.159);
        \node[below, font=\tiny] at (2.287,-0.05) {6};
        \draw[feBlue, line width=0.30pt] (2.669,0.626) -- (2.669,1.077);
        \draw[feBlue, line width=0.30pt] (2.669,1.700) -- (2.669,2.055);
        \draw[feBlue, line width=0.30pt] (2.569,0.626) -- (2.769,0.626);
        \draw[feBlue, line width=0.30pt] (2.569,2.055) -- (2.769,2.055);
        \filldraw[fill=feBox, draw=feBlue, line width=0.42pt] (2.569,1.077) rectangle (2.769,1.700);
        \draw[red!72!black, line width=0.38pt] (2.569,1.380) -- (2.769,1.380);
        \node[below, font=\tiny] at (2.669,-0.05) {7};
        \draw[feBlue, line width=0.30pt] (3.050,1.219) -- (3.050,1.301);
        \draw[feBlue, line width=0.30pt] (3.050,1.761) -- (3.050,1.829);
        \draw[feBlue, line width=0.30pt] (2.950,1.219) -- (3.150,1.219);
        \draw[feBlue, line width=0.30pt] (2.950,1.829) -- (3.150,1.829);
        \filldraw[fill=feBox, draw=feBlue, line width=0.42pt] (2.950,1.301) rectangle (3.150,1.761);
        \draw[red!72!black, line width=0.38pt] (2.950,1.533) -- (3.150,1.533);
        \node[below, font=\tiny] at (3.050,-0.05) {8};
        \draw[black!78, line width=0.52pt] (0.000,0.916) -- (0.381,1.031) -- (0.762,1.125) -- (1.144,1.171) -- (1.525,1.201) -- (1.906,1.178) -- (2.287,1.087) -- (2.669,1.379) -- (3.050,1.529);
        \fill[black!78] (0.000,0.916) circle (0.48pt);
        \fill[black!78] (0.381,1.031) circle (0.48pt);
        \fill[black!78] (0.762,1.125) circle (0.48pt);
        \fill[black!78] (1.144,1.171) circle (0.48pt);
        \fill[black!78] (1.525,1.201) circle (0.48pt);
        \fill[black!78] (1.906,1.178) circle (0.48pt);
        \fill[black!78] (2.287,1.087) circle (0.48pt);
        \fill[black!78] (2.669,1.379) circle (0.48pt);
        \fill[black!78] (3.050,1.529) circle (0.48pt);
        \end{scope}
        \node[font=\tiny] at (1.525,-0.43) {Generation};
        \node[font=\scriptsize, anchor=west] at (0,2.270) {Right: Generation distribution};
    \end{scope}
    \end{tikzpicture}%
    }
    \caption{Evolution stability of FE-alpha on CSI300. Left: mean test-set $|\mathrm{fitness}|$ trajectory across 20 initial seed programs, with shading indicating across-seed dispersion. Right: generation-level test-set $|\mathrm{fitness}|$ distribution across observed nodes.}
    \label{fig:fe_alpha_20_iteration_stability}
\end{figure}

\subsection{Token Efficiency and Executability}
Tab.~\ref{tab:cost} reports system-level efficiency under our implementation for three agent-based frameworks running 200 iterations. Baselines are evaluated using their official public implementations under identical hardware. The primary scarce resources are 
LLM/API calls and wall-clock time. FE reduces pressure on these resources by separating LLM-guided logic mutation from local numerical instantiation: the 20 Bayesian trials per macro mutation are executed without extra LLM calls and are included in the reported wall-clock time. 
The efficiency gains stem from two complementary sources: (1) the architectural separation that eliminates redundant LLM calls for parameter search and expression-to-code conversion, and (2) system-level implementation choices that minimize local evaluation latency. Both are integral to the FE framework. 
FE also achieves the highest executable ratio (99\%), reflecting the benefit of evolving directly within a validated code contract. 
\begin{table}[t]
\scriptsize
\centering
\caption{System runtime and executability under the 200-iteration setting. "Debug" is the ratio of API calls used for code debugging.}
\label{tab:cost}
\begin{adjustbox}{max width=\columnwidth}
\begin{tabular}{ccccccc}
\toprule
Methods & Search Space & Run Space  & Cost(\$)  & Time(h) & Executable Ratio & Debug
Ratio\\
\midrule
RD-Agent      & symbolic & code & 16.91 & 48.0 & 96\%            & 68\%       \\
AlphaAgent     & symbolic & code & 11.61 & 9.7  & 93\%            & 51\%       \\
FactorEngine      & code     & code & 12.01 & 0.5  & 99\%            & 32\%      \\
\bottomrule
\end{tabular}
\end{adjustbox}
\end{table}

\subsection{Ablation Study}

\begin{figure}[t]
    \centering
    \makebox[\linewidth][c]{%
    \begin{minipage}[t]{0.525\linewidth}
        \centering
        \includegraphics[width=\linewidth]{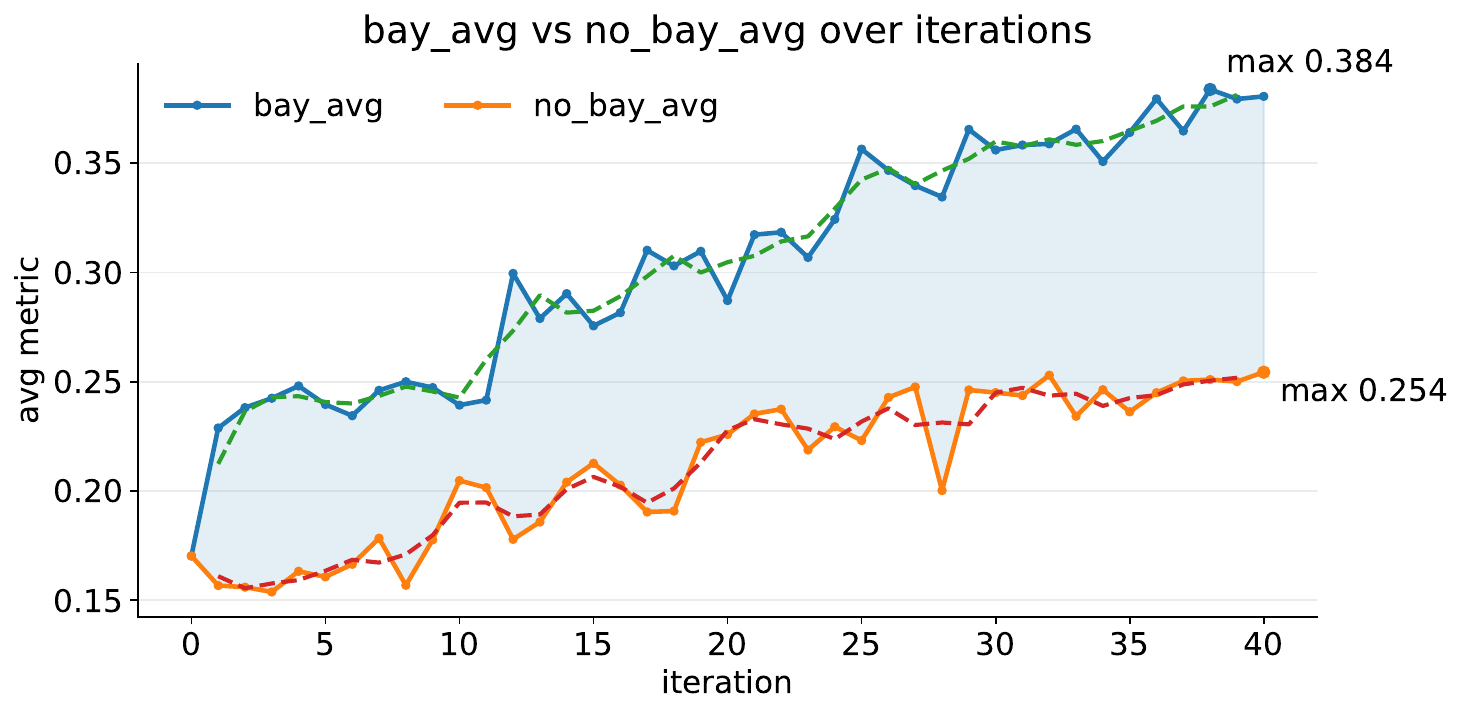}
    \end{minipage}%
    \hspace{-0.035\linewidth}%
    \begin{minipage}[t]{0.50\linewidth}
        \raggedright
            \begin{overpic}[width=0.82\linewidth,grid=false,trim=0 12pt 0 0,clip]{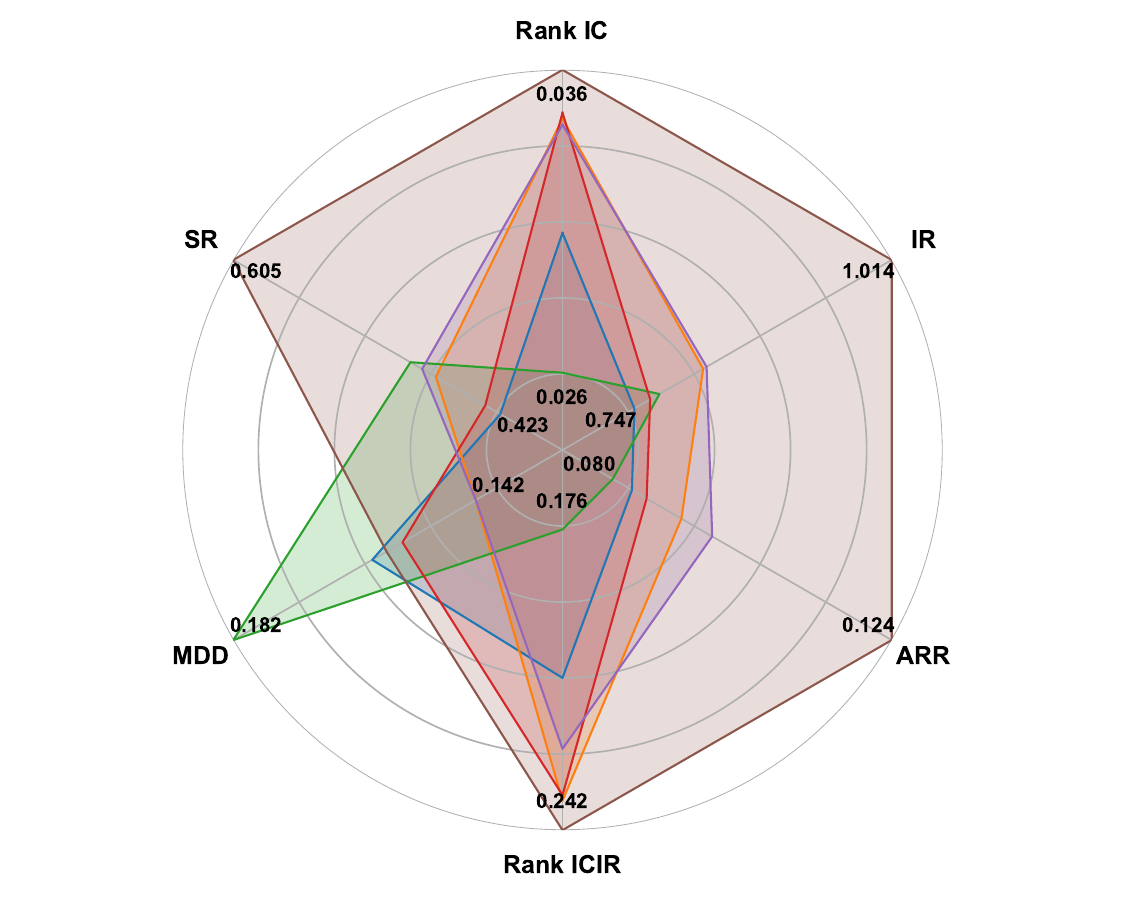}
                \put(83,54){%
                    \scalebox{0.78}{%
                    \begin{minipage}{0.36\linewidth}
                    \tiny
                    \definecolor{c1}{HTML}{1f77b4}
                    \definecolor{c2}{HTML}{2ba02b}
                    \definecolor{c3}{HTML}{9467bd}
                    \definecolor{c4}{HTML}{ff7f0f}
                    \definecolor{c5}{HTML}{d62727}
                    \definecolor{c6}{HTML}{8c564c}
                    \begin{tabular}{@{}l@{}}
                        \raisebox{0.7ex}{\textcolor{c1}{\rule{0.60em}{0.40pt}}}~RD-Agent-flash \\
                        \raisebox{0.7ex}{\textcolor{c2}{\rule{0.60em}{0.40pt}}}~AlphaAgent-flash \\
                        \raisebox{0.7ex}{\textcolor{c3}{\rule{0.60em}{0.40pt}}}~FE-alpha-flash \\
                        \raisebox{0.7ex}{\textcolor{c4}{\rule{0.60em}{0.40pt}}}~RD-Agent-gpt4o \\
                        \raisebox{0.7ex}{\textcolor{c5}{\rule{0.60em}{0.40pt}}}~AlphaAgent-gpt4o \\
                        \raisebox{0.7ex}{\textcolor{c6}{\rule{0.60em}{0.40pt}}}~FE-alpha-gpt4o \\
                    \end{tabular}
                    \end{minipage}
                    }
                }
            \end{overpic}
    \end{minipage}%
    }
    \caption{\textbf{Left}: Effect of Bayesian micro-instantiation. \texttt{bay\_avg} uses local Bayesian instantiation, while \texttt{no\_bay\_avg} uses fixed/default parameters. \textbf{Right}: Backbone robustness across GPT-4o and Gemini-2.5-flash-lite.}
    \label{fig:bay+back}
\end{figure}

\noindent \textbf{Bayesian Micro-Instantiation vs. Fixed Instantiation.}
We ablated the micro-level instantiation mechanism under the same macro-level 
evolution budget (40 iterations): \textbf{w/ Bayes} applied local Bayesian instantiation to each LLM-proposed factor logic family before evaluation, while \textbf{w/o Bayes} used fixed/default parameters. 
The left panel of Fig.~\ref{fig:bay+back} shows that Bayesian search improves both \emph{final quality} and \emph{search speed}. 
Concretely, the best evolved program after 40 iterations was substantially higher with Bayesian instantiation (about 0.38 vs.\ 0.25). Moreover, the improvement trajectory was consistently steeper, suggesting that validation-supported numerical instantiation provides a stronger and less noisy fitness signal for promoting promising program logic.

\noindent \textbf{Backbone Ablation.}
We conducted experiments using three agent-based frameworks with the Gemini-2.5-Flash-Lite\footnote{https://ai.google.
dev/gemini-api/docs/models} and GPT-4o\footnote{https://openai.com/index/gpt-4o-system-card/} for backbone ablation on the CSI300 market data. As shown in the right panel of Fig.~\ref{fig:bay+back}, all frameworks evolved for 200 iterations. 
Since GPT-4o generally exhibits stronger reasoning capabilities, the alpha factors generated by GPT-4o exhibited stronger predictive and portfolio performance.
Among the six setups, FE-alpha with GPT-4o showed the best results in Rank IC, Rank ICIR, AR, IR, and SR.

\noindent \textbf{Configuration Ablation.}
We further evaluate CoE prompts and multi-island evolution as search configurations that complement the core macro-micro mechanism.
Tab.~\ref{tab:parameter_ablation} evaluates their effect under different numbers of initial factors. Across both 5-alpha and 10-alpha settings, two-island configurations generally improve portfolio metrics relative to one-island configurations, suggesting that island diversification helps stabilize exploration. CoE prompts also improve several predictive and portfolio metrics, indicating that trajectory-conditioned context provides useful guidance.

\begin{table}[t]
\centering
\scriptsize
\caption{Comparisons of FE-alpha under different prompt settings, island configurations, and numbers of initial factors.}
\label{tab:parameter_ablation}
\begin{tabular}{l|ll|lll}
\toprule
\multicolumn{1}{c}{config} & \multicolumn{1}{c}{RIC} & \multicolumn{1}{c}{RICIR} & \multicolumn{1}{c}{AR} & \multicolumn{1}{c}{IR} & \multicolumn{1}{c}{|MDD|} \\
\midrule
5alpha,1island,CoE     & 0.0325 & 0.2165 & 0.0728 & 0.6737 & 0.1678 \\
5alpha,1island,top-k  & 0.0319 & 0.2125 & 0.0696 & 0.6428 & 0.1626 \\
5alpha,2island,CoE     & \textbf{0.0346} & \textbf{0.2308} & \textbf{0.0888} & \textbf{0.7886} & 0.1691 \\
5alpha,2island,top-k  & 0.0332 & 0.2189 & 0.0775 & 0.7085 & \textbf{0.1614} \\
\midrule
10alpha,1island,CoE    & 0.0344 & 0.2358 & 0.0782 & 0.7079 & 0.1787 \\
10alpha,1island,top-k & 0.0341 & 0.2266 & 0.0761 & 0.6944 & 0.1673 \\
10alpha,2island,CoE    & 0.0344 & \textbf{0.2360} & \textbf{0.0943} & \textbf{0.8241} & \textbf{0.1507} \\
10alpha,2island,top-k & \textbf{0.0353} & 0.2328 & 0.0839 & 0.7648 & 0.1708 \\
\bottomrule
\end{tabular}
\end{table}

\subsection{Case Study}
\label{sec:case_study}

We provide an example from the \emph{Abandoned Baby} report seed to illustrate how FE 
evolves a continuous, executable, and auditable factor from a report over 40 evolution iterations. 
The root program and the representative evolved node below are both taken from the same evolution tree. The root node has a fitness of $0.1006$, while the selected evolved node, found at
  iteration 40 and generation depth 4, reaches $0.3935$. Due to space, the listings show only factor-logic-related content and omit non-essential pre- and post-processing steps. Both programs still follow the same executable output contract.

\begin{compactpromptbox}{Core Root Program Excerpt}
\begin{lstlisting}[basicstyle=\ttfamily\fontsize{6.8}{7.1}\selectfont\color{black!92},caption={Root Abandoned-Baby program excerpt.},label={lst:ab_root_factor}]
def trend_factor(pricing_data: pl.DataFrame, parameters):
    # Omitted: imports, I/O preprocess, Parameter Definition
    # Root: direct three-candle continuous pattern scoring.
    df_pl = df_pl.with_columns([
        pl.col(c).shift(1).over('instrument').alias(f'{c}_lag1')
        for c in ['open', 'close', 'high', 'low'] ] + 
        [ pl.col(c).shift(2).over('instrument').alias(f'{c}_lag2')
        for c in ['open', 'close', 'high', 'low'] ])
    r2 = pl.col('high_lag2') - pl.col('low_lag2') + epsilon
    bull = (pl.col('close_lag2') - pl.col('open_lag2')) / r2
    r1 = pl.col('high_lag1') - pl.col('low_lag1') + epsilon
    body = (pl.col('open_lag1') - pl.col('close_lag1')).abs() / r1
    doji = (1.0 - body / body_threshold).clip(0.0, 1.0)
    gup = ( (pl.col('low_lag1') - pl.col('high_lag2')) /
        (pl.col('high_lag2') + epsilon) ).clip(-0.5, 0.5) + 0.5
    rt = pl.col('high') - pl.col('low') + epsilon
    bear = (pl.col('open') - pl.col('close')) / rt
    gdn = ( (pl.col('low_lag1') - pl.col('high')) /
        (pl.col('low_lag1') + epsilon) ).clip(-0.5, 0.5) + 0.5
    # Rolling context keeps this as executable logic.
    vavg = pl.col('volume').rolling_mean(window_size=vol_lookback, 
        min_periods=1).over('instrument')
    vol = ((pl.col('volume') / (vavg + epsilon)) /
        vol_multiplier).clip(0.0, 2.0)
    pq = pl.col('close').rolling_quantile( quantile=price_quantile,
        window_size=price_lookback, min_periods=1 ).over('instrument')
    px = (pl.col('close') / (pq + epsilon)).clip(0.5, 1.5)
    F = -( 0.15 * bull + 0.20 * doji + 0.15 * gup +
        0.20 * bear + 0.15 * gdn + 0.075 * vol + 0.075 * px)
    df_factor = df_pl.with_columns(F.cast(pl.Float64).alias('TrendFactor'))
    # Omitted: select columns, filter invalid values.
    return df_factor
\end{lstlisting}
\end{compactpromptbox}

\begin{compactpromptbox}{Core Evolved Program Excerpt (Iteration 40)}
\begin{lstlisting}[basicstyle=\ttfamily\fontsize{6.8}{7.1}\selectfont\color{black!92},caption={Evolved Abandoned-Baby program excerpt.},label={lst:ab_evolved_factor}]
def trend_factor(pricing_data: pl.DataFrame, parameters):
    # Omitted: imports, I/O preprocess, Parameter Definition
    # Evolved: adds ROC context, ATR scaling, exponential doji, range position.
    df_pl = df_pl.with_columns([
        pl.col('close').shift(1).over('instrument').alias('close_lag1') ])
    tr = pl.max_horizontal( pl.col('high') - pl.col('low'),
        (pl.col('high') - pl.col('close_lag1')).abs(),
        (pl.col('low') - pl.col('close_lag1')).abs()    )
    df_pl = df_pl.with_columns( tr.ewm_mean(span=atr_lookback,
        min_periods=atr_lookback).over('instrument').alias('atr') )
    df_pl = df_pl.with_columns([pl.col(c).shift(1).over('instrument')
        .alias(f'{c}_lag1')
        for c in ['open', 'close', 'high', 'low', 'atr'] ] + 
        [ pl.col(c).shift(2).over('instrument').alias(f'{c}_lag2')
        for c in ['open', 'close', 'high', 'low', 'atr'] ] )
    p0 = pl.col('close_lag2').shift(roc_lookback).over('instrument')
    mom = ((pl.col('close_lag2') - p0) / (p0 + epsilon)).clip(0, 1.0)
    r1 = pl.col('high_lag1') - pl.col('low_lag1') + epsilon
    body = (pl.col('open_lag1') - pl.col('close_lag1')).abs() / r1
    doji = (-doji_exp_decay * body).exp()
    gup = ((pl.col('low_lag1') - pl.col('high_lag2')) / 
        (pl.col('atr_lag2') + epsilon)).clip(0, 1.5)
    bear = ((pl.col('open') - pl.col('close')) / (pl.col('atr') +   
        epsilon)).clip(0, None)
    gdn = ((pl.col('low_lag1') - pl.col('high')) / (pl.col('atr_lag1') + 
        epsilon)).clip(0, 1.5)
    vavg = pl.col('volume').rolling_mean(window_size=vol_lookback, 
        min_periods=1).over('instrument')
    vol = ((pl.col('volume') / (vavg + epsilon)) / 
        vol_multiplier).clip(0.0, 2.0)
    lo = pl.col('low').rolling_min(window_size=price_lookback, 
        min_periods=price_lookback // 2).over('instrument')
    hi = pl.col('high').rolling_max(window_size=price_lookback, 
        min_periods=price_lookback // 2).over('instrument')
    px = ((pl.col('close') - lo) / (hi - lo + epsilon)).clip(0.0, 1.0)
    F = -(  0.25 * mom + 0.20 * doji + 0.15 * gup +
        0.10 * bear + 0.15 * gdn + 0.075 * vol + 0.075 * px )
    df_factor = df_pl.with_columns(F.cast(pl.Float64).alias('TrendFactor'))
    # Omitted: select columns, filter invalid values.
    return df_factor
\end{lstlisting}
\end{compactpromptbox}

This example presents a report-inspired alpha whose evolved program acts as a structured candlestick-pattern interpreter, combining lagged candle semantics, true range and EWM-based ATR volatility normalization, exponential doji-quality scoring, ATR-scaled gap strength, ROC-based prior momentum, and volume/price-position conditioning. The factor remains continuous and executable under the seed program’s output contract.
Fig.~\ref{fig:ab_generation_bubble} shows Bayesian micro-instantiation trials for the selected generation-4 evolved program in Listing~\ref{lst:ab_evolved_factor}: 15 out of 20 trials produced valid scores, while invalid trials were discarded due to execution and contract checks. 

\begin{figure}[t]
\centering
\begin{tikzpicture}[x=0.8cm,y=0.8cm]
    \definecolor{mainblue}{HTML}{1f5fa8}
    \definecolor{softblue}{HTML}{d9e8f7}
    \definecolor{accentred}{HTML}{b03a2e}
    \draw[->, line width=0.45pt] (0.25,0.35) -- (6.0,0.35) node[right, font=\scriptsize] {Generation};
    \draw[->, line width=0.45pt] (0.42,0.18) -- (0.42,3.78) node[above, font=\scriptsize] {Mean $|\mathrm{fitness}|$};
    \foreach \x/\lab in {0.65/0,1.85/1,3.0/2,4.25/3,5.5/4} {
        \draw[black!40] (\x,0.32) -- (\x,0.38);
        \node[font=\scriptsize] at (\x,0.12) {\lab};
    }
    \foreach \y/\lab in {0.49/0.10,1.06/0.15,1.63/0.20,2.20/0.25,2.77/0.30,3.34/0.35} {
        \draw[black!14, dashed, line width=0.22pt] (0.42,\y) -- (5.75,\y);
        \node[left, font=\scriptsize] at (0.34,\y) {\lab};
    }
    \draw[mainblue, line width=0.85pt]
        (0.65,0.49) -- (1.85,0.72) -- (3.0,2.61) -- (4.25,2.91) -- (5.5,3.23);
    \foreach \x/\y/\r in {
        0.65/0.49/0.09,
        1.85/0.72/0.34,
        3.0/2.61/0.31,
        4.25/2.91/0.18,
        5.5/3.23/0.15
    } {
        \filldraw[fill=softblue, draw=mainblue, line width=0.6pt] (\x,\y) circle (\r);
    }
    \node[font=\tiny, anchor=south west] at (0.72,0.55) {$n=1$};
    \node[font=\tiny, anchor=south] at (1.85,1.08) {$n=9$};
    \node[font=\tiny, anchor=south] at (3.0,2.94) {$n=16$};
    \node[font=\tiny, anchor=south] at (4.25,3.12) {$n=8$};
    \node[font=\tiny, anchor=south] at (5.5,3.40) {$n=4$};
    \draw[black!35, rounded corners=1pt, line width=0.28pt, fill=white, fill opacity=0.97]
        (2.67,0.58) rectangle (5.86,1.99);
    \node[font=\tiny, anchor=west] at (2.89,1.86) {Bayesian trials};
    \draw[black!40, line width=0.25pt] (3.14,0.92) -- (5.56,0.92);
    \draw[black!40, line width=0.25pt] (3.14,0.88) -- (3.14,1.59);
    \foreach \y/\lab in {0.92/0.36,1.17/0.37,1.41/0.38} {
        \draw[black!14, line width=0.2pt] (3.14,\y) -- (5.56,\y);
        \node[font=\tiny, rotate=25, anchor=east] at (3.19,\y-0.01) {\lab};
    }
    \draw[accentred, dashed, line width=0.35pt] (3.14,1.25) -- (5.56,1.25);
    \node[font=\tiny, accentred, anchor=north] at (5.30,1.23) {mean};
    \foreach \x/\y in {
        3.17/0.93,3.32/0.95,3.47/1.05,3.62/1.06,
        3.77/1.19,3.92/1.21,4.07/1.22,4.22/1.22,
        4.37/1.31,4.52/1.32,4.67/1.33,4.82/1.35,
        4.97/1.38,5.12/1.40,5.27/1.51} {
        \filldraw[fill=mainblue, draw=white, line width=0.15pt] (\x,\y) circle (1.12pt);
    }
    \filldraw[fill=accentred, draw=white, line width=0.15pt] (5.42,1.52) circle (1.45pt);
    \node[font=\tiny, accentred, anchor=south] at (5.42,1.56) {best};
\end{tikzpicture}
\caption{\textbf{Abandoned-Baby evolution quality.} Bubbles show mean training $|\mathrm{fitness}|$ by generation, with radius encoding within-generation standard deviation. The inset reports valid Bayesian-instantiation scores for the selected generation-4 node.}
\label{fig:ab_generation_bubble}
\end{figure}

\section{Conclusion}
We presented \emph{FactorEngine}, a program-level alpha factor mining framework for discovering \emph{executable and auditable} factors while keeping the evolution tractable. FE departs from prior symbolic expression search by representing factors as execution-constrained programs and separating logic evolution from market-conditioned instantiation.
FE further introduces a knowledge-infused module that transforms financial reports into executable seeds via a closed-loop extraction--verification--generation pipeline. CoE prompts and island search act as supporting mechanisms for trajectory-conditioned context and diversity-preserving exploration.
Overall, FE provides an empirical framework for factor discovery in discrete program spaces, where LLM-guided structural mutation is paired with local Bayesian micro-instantiation under validation feedback.
Future work includes extending to richer data modalities, improving robustness under distribution shift and enabling the LLM to actively interrogate market data.
\bibliographystyle{IEEEtran}
\bibliography{reference}

\end{document}